\documentclass{article}
\usepackage{spconf,amsmath,graphicx,hyperref}

\usepackage{multirow}
\usepackage{booktabs}
\usepackage[table]{xcolor}
\usepackage{amssymb}
\usepackage{arydshln}


\title{Hashing-Baseline: Rethinking hashing \\in the age of pretrained models}
\name{Ilyass Moummad*$^{1}$ \quad Kawtar Zaher*$^{1,2}$ \quad Lukas Rauch$^{3}$ \quad Alexis Joly$^{1}$\thanks{*Equal contributions.}}
\address{
  $^{1}$INRIA, LIRMM, Université de Montpellier, France \\
  $^{2}$Institut National de l’Audiovisuel, France \\
  $^{3}$University of Kassel, Germany
}

\begin{document}
%
\maketitle
\begin{abstract}
Information retrieval with compact binary codes, also referred to as \textit{hashing}, is crucial for scalable fast search applications, yet state-of-the-art hashing methods require expensive, scenario-specific training. In this work, we introduce \textit{Hashing-Baseline}, a strong training-free hashing method leveraging powerful pre-trained encoders that produce rich embeddings. We revisit classical, training-free hashing techniques—principal component analysis, random orthogonal projection, and threshold binarization—to produce a strong baseline for hashing. Our approach combines these techniques with frozen embeddings from state-of-the-art vision and audio encoders to yield competitive retrieval performance without any additional learning or fine-tuning. To demonstrate the generality and effectiveness of this approach, we evaluate it on standard image retrieval benchmarks as well as a newly introduced benchmark for audio hashing.\footnote{Code is released at: \url{https://github.com/ilyassmoummad/hashing-baseline}}
\end{abstract}


\section{Introduction}
\label{sec:intro}

\begin{figure*}[!h]
  \centering
  \includegraphics[width=0.85\textwidth]{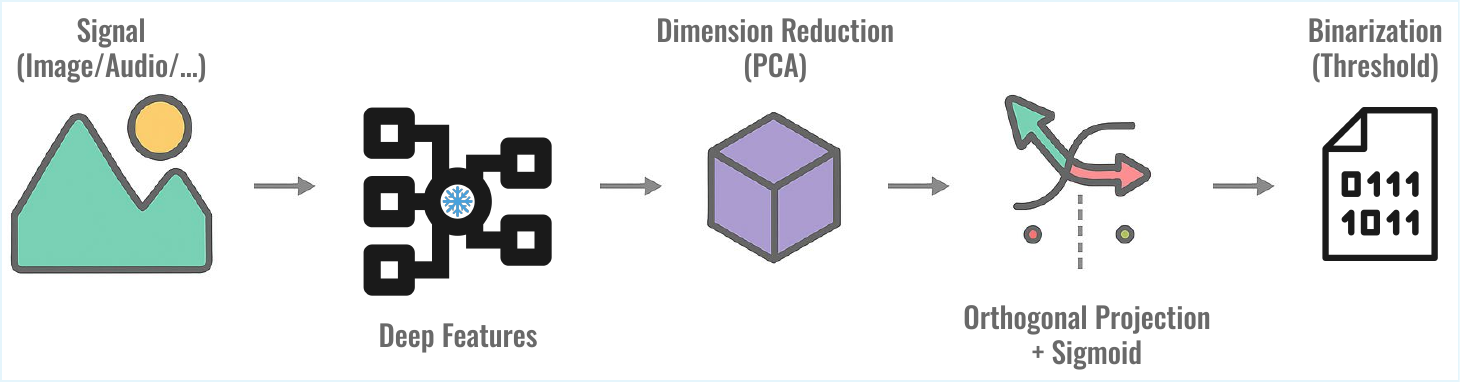}
  \caption{Overview of \textit{Hashing-Baseline}: Features are extracted using a frozen pre-trained model, then reduced to the target bit lengths via PCA. The reduced features are orthogonally projected and binarized using a sigmoid function followed by a threshold to generate compact binary codes.}
  \label{fig:pipeline}
\end{figure*}

Fast and accurate retrieval using binary embeddings is essential for large-scale search. Traditional hashing methods rely on handcrafted descriptors to generate compact binary codes, while more recent deep hashing techniques—both supervised and unsupervised—usually require training models from scratch. This training process is often computationally expensive and time-consuming~\cite{deephashing}. Moreover, these methods generally lack flexibility, as they must be retrained for each specific scenario, such as different code lengths or different datasets, limiting their scalability and generalization.

Concurrently, the emergence of foundation models has revolutionized data representations by producing powerful embeddings in abstract latent spaces learned from vast and diverse datasets~\cite{bommasani2021opportunities}. These representations capture rich semantic information, providing a strong starting point for various downstream tasks. This naturally motivates the question: can we rethink hashing by leveraging these pretrained embeddings directly, rather than investing in costly, scenario-specific training of hashing networks?

To address this, we introduce \textit{Hashing-Baseline}, a training-free method that re-examines classical hashing techniques, namely Principal Component Analysis (PCA), random orthogonal projections~\cite{achlioptas2003database}, and binarization via thresholding. When these techniques are applied to embeddings from pre-trained encoders, their combination consistently yields competitive, and sometimes state-of-the-art, performances, all achieved without the need for further learning.

Motivated by the increasing importance of audio retrieval, we also establish the first comprehensive benchmark dedicated to audio hashing. In contrast to classification tasks—where modern pre-trained audio models often achieve very high accuracy~\cite{icmeaudio}, making further progress difficult—retrieval provides a more demanding and discriminative test of audio understanding. Even when relevance is defined at the class level, successful retrieval requires ranking all relevant items ahead of all irrelevant ones, placing stricter constraints on embedding space structure than classification, which only needs the top label to be correct. Our benchmark, covering diverse audio types including music genres, speech emotions, human vocalizations, and environmental sounds, together with \textit{Hashing-Baseline}, provides a comprehensive and challenging testbed for audio hashing.


\vspace{1em}
In summary, our contributions are twofold:
\begin{itemize}
    \item We propose \textit{Hashing-Baseline}, a strong training-free method that leverages pre-trained model embeddings for efficient retrieval;
    \item We establish a new benchmark for audio hashing, spanning diverse audio domains (music genres, speech emotions, human vocalizations, and environmental sounds).
\end{itemize}

\section{Related Work}
\label{sec:related}

\textbf{Pre-trained models} have significantly advanced deep learning by providing robust, pre-trained representations applicable across a wide range of tasks~\cite{dinov2, dasheng}. Trained on extensive and diverse datasets, these models capture rich semantic information and have demonstrated exceptional performance in numerous downstream applications~\cite{bommasani2021opportunities}. Their emergence reduces the reliance on task-specific training, making it possible to design more efficient and scalable solutions directly on top of frozen embeddings.

\noindent \textbf{Hashing} enables fast similarity search in large-scale retrieval systems by encoding data into compact binary representations. Traditional hashing methods apply predefined or data-driven transformations—such as random projections or spectral analysis—to preserve similarity in high-dimensional spaces. While efficient, these approaches struggle to capture complex semantic relationships from raw data~\cite{trasdhashing}. Deep learning-based hashing instead employs neural networks to learn data-dependent hash functions. Supervised variants preserve semantic similarity using labeled data, while unsupervised variants rely on intrinsic data properties~\cite{deephashing}. Although deep hashing often produces more discriminative codes than traditional methods, it typically requires computationally expensive training tailored to specific datasets and code lengths, limiting its generalization and scalability.

Our proposed method, \textit{Hashing-Baseline}, bridges these two lines of research by leveraging the strong representational power of pre-trained models together with classical, training-free hashing techniques. Unlike traditional hashing, \textit{Hashing-Baseline} operates on rich pretrained embeddings rather than raw features, and unlike deep hashing, it requires no additional training. This combination yields a simple, scalable, and surprisingly competitive baseline for both image and audio retrieval tasks.

\section{Hashing-Baseline Method}
\label{sec:method}

\begin{figure*}[!t]
  \centering
  \includegraphics[width=\textwidth]{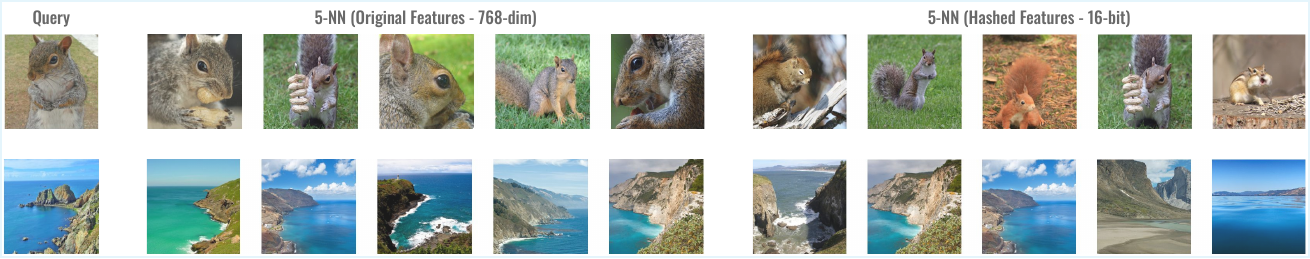}
  \caption{Retrieval examples on Flickr25K, showing the nearest neighbors using SimDINO features and their 16-bit hashed codes.}
  \label{fig:retrieval}
\end{figure*}

We denote by $\mathbf{s} \in \mathbb{R}^T$ a signal from any modality (e.g., an image or audio clip). 
A pre-trained encoder $f_{\phi}(\cdot)$ maps $\mathbf{s}$ to a $d$-dimensional feature vector:
\begin{equation}
    \mathbf{x} = f_{\phi}(\mathbf{s}) \in \mathbb{R}^d.
\end{equation}

\textit{Hashing-Baseline} then consists of three simple steps: (i) dimensionality reduction with PCA, (ii) random orthogonal projection, and (iii) binarization with asymmetric Hamming retrieval. 

\subsection{Dimensionality Reduction via PCA}
Let $\mathbf{X} = \{\mathbf{x}_i\}_{i=1}^n \in \mathbb{R}^{n \times d}$ denote a training set of normalized embeddings. 
We compute a truncated SVD:
\begin{equation}
    \mathbf{X} \approx \mathbf{U} \, \mathrm{diag}(\mathbf{S}) \, \mathbf{V}^{\top}, 
\end{equation}
where the columns of $\mathbf{V}$ are the top-$k$ principal directions. 
A feature $\mathbf{x}$ is then projected into the reduced $k$-dimensional space:
\begin{equation}
    \mathbf{z} = \mathbf{V}^{\top}\mathbf{x} \in \mathbb{R}^k.
\end{equation}

\subsection{Random Orthogonal Projection and Binarization}
A random orthogonal matrix $\mathbf{R} \in \mathbb{R}^{k \times k}$ is generated by sampling a Gaussian matrix followed by QR decomposition. 
The reduced vector is transformed as
\begin{equation}
    \mathbf{u} = \mathbf{R}\mathbf{z}.
\end{equation}
We then apply an entry-wise sigmoid to obtain bit probabilities:
\begin{equation}
    \mathbf{p} = \sigma(\mathbf{u}) \in [0,1]^k,
\end{equation}
and generate binary codes $\mathbf{b} \in \{0,1\}^k$ for database items by thresholding $\mathbf{p}$ at $0.5$.

\subsection{Asymmetric Hamming Retrieval}
For a query embedding $\mathbf{x}_q$, we compute
\begin{equation}
    \mathbf{p}_q = \sigma(\mathbf{R}\mathbf{V}^{\top}\mathbf{x}_q).
\end{equation}
Rather than binarizing the query, we use \emph{asymmetric hamming retrieval}~\cite{asymham}, which compares $\mathbf{p}_q$ to binary database codes $\mathbf{b}_i$ via:
\begin{equation}
    \mathrm{sim}(\mathbf{x}_q, \mathbf{b}_i) = - \sum_{j=1}^k \left| b_{i,j} - p_{q,j} \right|.
\end{equation}
Asymmetric Hamming allows to reduce quantization loss on the query-level, and thus improves retrieval accuracy.

\subsection{Theoretical Motivation}
\textit{Hashing-Baseline} is grounded in classical results: (i) PCA reduces information redundancy and noise in the pre-trained embeddings, and retains only the most informative dimensions, after which random projections preserve geometry in the reduced space with high probability, as ensured by the Johnson–Lindenstrauss lemma~\cite{johnson}; 
(ii) random orthogonal projections redistribute the concentrated variance across bits; 
(iii) random hyperplane hashing~\cite{tessellations} links Hamming distance to angular similarity.

Together, these properties suggest that combining pre-trained model embeddings with classical hashing can provide strong retrieval performance without any additional learning.

\section{Experiments}
\label{sec:experiments}

We evaluate the proposed \textit{Hashing-Baseline} on both image and audio retrieval benchmarks, measuring mean Average Precision (mAP). 
 
For images, we follow standard practice and report mAP@1000 on CIFAR-10 and mAP@5000 on Flickr25K, COCO, and NUS-WIDE~\cite{fsch}.  
For audio, mAP is computed over all database items.

\begin{table*}[!h]
\centering
\caption{Image retrieval with different ViT-Base models. 
\textit{Orig}: full embeddings; 
\textit{Float}: PCA-reduced features; 
\textit{Binary}: random-projected codes (mean$\pm$std over 10 runs).
SOTA indicates the state-of-the-art of unsupervised hashing.}
\label{tab:image_retrieval}
\resizebox{\textwidth}{!}{
\begin{tabular}{clcccccccccccccccc}
\toprule
\multirow{2}{*}{\textbf{Model}} & \multirow{2}{*}{\textbf{Features}} 
& \multicolumn{4}{c}{\textbf{CIFAR10}} 
& \multicolumn{4}{c}{\textbf{FLICKR25K}} 
& \multicolumn{4}{c}{\textbf{COCO}} 
& \multicolumn{4}{c}{\textbf{NUS-WIDE}} \\
\cmidrule(lr){3-6} \cmidrule(lr){7-10} \cmidrule(lr){11-14} \cmidrule(lr){15-18}
& & Orig & 16 & 32 & 64 
  & Orig & 16 & 32 & 64 
  & Orig & 16 & 32 & 64 
  & Orig & 16 & 32 & 64 \\
\midrule

\multirow{1}{*}{\textcolor{gray}{\shortstack{\textbf{SOTA}}}}
& Binary   & -- & \textcolor{gray}{87.6$^{\cite{fsch}}$} & \textcolor{gray}{91.2$^{\cite{fsch}}$} & \textcolor{gray}{92.6$^{\cite{fsch}}$} 
           & -- & \textcolor{gray}{81.8$^{\cite{harr}}$} & \textcolor{gray}{83.8$^{\cite{fsch}}$} & \textcolor{gray}{84.9$^{\cite{fsch}}$} 
           & -- & \textcolor{gray}{76.0$^{\cite{fsch}}$} & \textcolor{gray}{78.9$^{\cite{harr}}$} & \textcolor{gray}{81.6$^{\cite{harr}}$} 
           & -- & \textcolor{gray}{81.2$^{\cite{fsch}}$} & \textcolor{gray}{83.2$^{\cite{fsch}}$} & \textcolor{gray}{84.4$^{\cite{fsch}}$} \\

\midrule

\multirow{2}{*}{\shortstack{DFN~\cite{dfn} \\ \footnotesize{(DFN-2B)}}}
& Float    & 93.3 & 94.6 & 94.4 & 94.22 
           & 80.7 & 83.7 & 83.9 & 83.6 
           & 85.3 & 77.1 & 82.3 & 85.3 
           & 83.2 & 81.9 & 83.1 & 83.2 \\
& Binary   & -- & 91.0\footnotesize{$\pm$0.7} & 91.8\footnotesize{$\pm$0.4} & 92.6\footnotesize{$\pm$0.2} 
           & -- & 80.2\footnotesize{$\pm$0.3} & 80.8\footnotesize{$\pm$0.3} & 81.1\footnotesize{$\pm$0.1} 
           & -- & 70.4\footnotesize{$\pm$0.6} & 77.4\footnotesize{$\pm$0.2} & 82.2\footnotesize{$\pm$0.1} 
           & -- & 76.7\footnotesize{$\pm$0.4} & 79.9\footnotesize{$\pm$0.2} & 81.1\footnotesize{$\pm$0.1} \\

\midrule

\multirow{2}{*}{\shortstack{DINOv2~\cite{dinov2} \\ \footnotesize{(LVD-142M)}}}
& Float    & 95.4 & 95.9 & 96.0 & 95.9 
           & 76.3 & 77.8 & 78.2 & 77.7 
           & 88.3 & 81.2 & 86.5 & 88.8 
           & 79.8 & 76.4 & 78.0 & 78.7 \\
& Binary   & -- & 93.4\footnotesize{$\pm$0.5} & 95.0\footnotesize{$\pm$0.1} & 94.7\footnotesize{$\pm$0.1} 
           & -- & 74.3\footnotesize{$\pm$0.2} & 74.9\footnotesize{$\pm$0.1} & 74.7\footnotesize{$\pm$0.1} 
           & -- & 74.9\footnotesize{$\pm$0.5} & 83.5\footnotesize{$\pm$0.2} & 86.7\footnotesize{$\pm$0.1} 
           & -- & 70.6\footnotesize{$\pm$0.5} & 73.8\footnotesize{$\pm$0.2} & 75.9\footnotesize{$\pm$0.1} \\

\midrule



\multirow{6}{*}{\shortstack{SimDINOv2~\cite{simdino} \\ \footnotesize{(IN-1K)}}}
& Float    & 89.6 & 90.8 & 91.1 & 91.13 
           & 81.1 & 81.6 & 81.6 & 81.4 
           & 87.4 & 82.7 & 86.0 & 87.3 
           & 84.3 & 83.2 & 83.7 & 83.6 \\
& Binary   & -- & 84.4\footnotesize{$\pm$0.4} & 86.4\footnotesize{$\pm$0.4} & 88.0\footnotesize{$\pm$0.2} 
           & -- & 77.2\footnotesize{$\pm$0.5} & 78.0\footnotesize{$\pm$0.1} & 78.6\footnotesize{$\pm$0.1} 
           & -- & 75.5\footnotesize{$\pm$0.4} & 81.8\footnotesize{$\pm$0.3} & 84.9\footnotesize{$\pm$0.1} 
           & -- & 78.0\footnotesize{$\pm$0.2} & 80.3\footnotesize{$\pm$0.2} & 81.5\footnotesize{$\pm$0.1} \\
\cmidrule(lr){2-18}
& \multicolumn{17}{l}{\textit{\textbf{Ablation Studies}}} \\
\cmidrule(lr){2-3}
& \multicolumn{17}{l}{\textit{Global PCA (IN-1K)}} \\
& Float    & 89.6 & 80.9 & 87.7 & 89.0 
           & 81.1 & 79.7 & 81.2 & 81.2 
           & 87.4 & 78.3 & 82.8 & 85.2 
           & 84.3 & 78.7 & 81.9 & 83.0 \\
& Binary   & -- & 66.0\footnotesize{$\pm$1.7} & 79.4\footnotesize{$\pm$0.8} & 84.0\footnotesize{$\pm$0.5} 
           & -- & 75.1\footnotesize{$\pm$0.9} & 77.3\footnotesize{$\pm$0.5} & 78.3\footnotesize{$\pm$0.2} 
           & -- & 69.2\footnotesize{$\pm$0.9} & 77.2\footnotesize{$\pm$0.3} & 81.8\footnotesize{$\pm$0.2} 
           & -- & 72.7\footnotesize{$\pm$0.6} & 77.7\footnotesize{$\pm$0.2} & 80.5\footnotesize{$\pm$0.2} \\
& \multicolumn{17}{l}{\textit{Global PCA without random orthogonal projection}} \\
& Binary   & -- & 65.9 & 77.9 & 81.5 
           & -- & 73.6 & 75.1 & 74.2 
           & -- & 68.9 & 76.6 & 81.1 
           & -- & 73.6 & 78.6 & 80.0 \\         
& \multicolumn{17}{l}{\textit{Random orthogonal projection without PCA}} \\
& Binary   & -- & 40.7\footnotesize{$\pm$2.8} & 59.3\footnotesize{$\pm$2.7} & 71.9\footnotesize{$\pm$1.4} 
           & -- & 61.1\footnotesize{$\pm$1.0} & 65.2\footnotesize{$\pm$1.4} & 69.1\footnotesize{$\pm$0.8}  
           & -- & 54.9\footnotesize{$\pm$1.2} & 68.0\footnotesize{$\pm$1.0} & 76.3\footnotesize{$\pm$0.5} 
           & -- & 55.4\footnotesize{$\pm$2.2} & 65.4\footnotesize{$\pm$1.5} & 73.7\footnotesize{$\pm$0.7} \\

\bottomrule
\end{tabular}
}
\end{table*}

\begin{table*}[!h]
\centering
\caption{Audio retrieval using the same protocol as in table~\ref{tab:image_retrieval}.}
\label{tab:audio_retrieval}
\resizebox{\textwidth}{!}{
\begin{tabular}{clcccccccccccccccc}
\toprule
\multirow{2}{*}{\textbf{Model}} & \multirow{2}{*}{\textbf{Features}} 
& \multicolumn{4}{c}{\textbf{GTZAN}} 
& \multicolumn{4}{c}{\textbf{ESC50}} 
& \multicolumn{4}{c}{\textbf{VocalSound}} 
& \multicolumn{4}{c}{\textbf{CREMA-D}} \\
\cmidrule(lr){3-6} \cmidrule(lr){7-10} \cmidrule(lr){11-14} \cmidrule(lr){15-18}
& & Orig & 16 & 32 & 64 & Orig & 16 & 32 & 64 & Orig & 16 & 32 & 64 & Orig & 16 & 32 & 64 \\
\midrule

\multirow{2}{*}{CLAP~\cite{laionclap}}
& Float  & 41.2 & 41.2 & 38.2 & 37.4 & 88.1 & 81.4 & 87.3 & 87.7 & 62.7 & 59.3 & 57.0 & 55.7 & 25.1 & 25.1 & 25.0 & 24.9 \\
& Binary & -- & 37.9\footnotesize{$\pm$1.0} & 39.3\footnotesize{$\pm$0.7} & 40.9\footnotesize{$\pm$0.3} 
         & -- & 70.0\footnotesize{$\pm$1.5} & 81.2\footnotesize{$\pm$0.5} & 84.7\footnotesize{$\pm$0.3} 
         & -- & 58.3\footnotesize{$\pm$0.9} & 60.3\footnotesize{$\pm$0.5} & 61.6\footnotesize{$\pm$0.2} 
         & -- & 24.0\footnotesize{$\pm$0.1} & 24.5\footnotesize{$\pm$0.2} & 24.9\footnotesize{$\pm$0.2} \\

\midrule

\multirow{2}{*}{Dasheng~\cite{dasheng}}
& Float  & 38.4 & 40.6 & 39.1 & 36.7 & 29.8 & 27.4 & 35.1 & 39.4 & 27.8 & 31.7 & 31.8 & 31.6 & 25.0 & 24.9 & 25.3 & 25.2 \\
& Binary & -- & 35.1\footnotesize{$\pm$1.0} & 36.8\footnotesize{$\pm$0.9} & 37.8\footnotesize{$\pm$0.5} 
         & -- & 19.0\footnotesize{$\pm$0.6} & 25.3\footnotesize{$\pm$0.4} & 29.3\footnotesize{$\pm$0.5} 
         & -- & 26.6\footnotesize{$\pm$0.4} & 27.5\footnotesize{$\pm$0.3} & 28.1\footnotesize{$\pm$0.2} 
         & -- & 23.9\footnotesize{$\pm$0.2} & 24.6\footnotesize{$\pm$0.1} & 25.0\footnotesize{$\pm$0.1} \\

\midrule

\multirow{2}{*}{CED~\cite{ced}}
& Float  & 51.5 & 53.7 & 50.0 & 48.3 & 82.7 & 50.0 & 72.8 & 83.2 & 60.2 & 58.7 & 58.5 & 58.5 & 19.3 & 20.6 & 20.6 & 20.7 \\
& Binary & -- & 51.4\footnotesize{$\pm$0.8} & 51.3\footnotesize{$\pm$0.5} & 52.4\footnotesize{$\pm$0.2} 
         & -- & 64.1\footnotesize{$\pm$2.3} & 79.5\footnotesize{$\pm$0.3} & 82.2\footnotesize{$\pm$0.3} 
         & -- & 57.3\footnotesize{$\pm$0.7} & 58.8\footnotesize{$\pm$0.5} & 60.2\footnotesize{$\pm$0.3} 
         & -- & 19.2\footnotesize{$\pm$0.1} & 19.4\footnotesize{$\pm$0.1} & 19.5\footnotesize{$\pm$0.1} \\

\bottomrule
\end{tabular}
}
\end{table*}

\subsection{Image Retrieval}
\label{subsec:image_retrieval}

We evaluate \textit{Hashing-Baseline} on three state-of-the-art ViT-Base vision encoders: \textbf{DFN}~\cite{dfn}, trained on 2B image-text pairs via contrastive learning; \textbf{DINOv2}~\cite{dinov2}, trained on a curated dataset of 142 million images; and \textbf{SimDINOv2}~\cite{simdino}, a DINOv2 variant trained with cosine similarity and coding-rate regularization on ImageNet-1K. For each model, we report retrieval results using embeddings reduced via PCA fitted on the training set of each downstream dataset, alongside the original embeddings and corresponding $d$-bit binary codes obtained via random orthogonal projection and sign thresholding ($d \in {16, 32, 64}$).

To further study the generalization of \textit{Hashing-Baseline}, we conduct ablation experiments on SimDINOv2 using a single PCA fitted on ImageNet-1K (global PCA) applied to all datasets, and additionally evaluate the effects of: (i) global PCA without random orthogonal projection, and (ii) random orthogonal projection applied without PCA. This setup allows us to isolate the contributions of PCA and orthogonal projection to retrieval performance. Results are summarized in Table~\ref{tab:image_retrieval}.



\noindent\textbf{Analysis.}  
Across all datasets, \textit{Hashing-Baseline} preserves much of the original embedding quality (especially at 64-bit). Surprisingly, even with only 16 bits, it reaches very high mAP on several benchmarks without any additional learning. The obtained results suggests that strong pre-trained model features contain a high degree of redundancy: while such redundancy may be beneficial during pre-training, it appears not strictly necessary for downstream retrieval tasks. Finally, our ablation study confirms that PCA and random orthogonal projection—two core components of \textit{Hashing-Baseline}—play complementary roles, as removing either significantly reduces performance.

\begin{table}[!h]
\centering
\caption{Audio datasets retrieval benchmark.}
\label{tab:audiodatasets}
\resizebox{\columnwidth}{!}{
\begin{tabular}{lcccc}
\toprule
Dataset & Train/Database & Validation & Query & Classes \\
\midrule
GTZAN~\cite{gtzan} & 400 & — & 599 & 10 \\
CREMA-D~\cite{cremad} & 5,210  & 1,116 & 1,116 & 6 \\
VocalSound~\cite{vocalsound} & 15,531 & 1,855 & 3,591 & 6 \\
ESC-50~\cite{esc50} & 400 & — & 1,600 & 50 \\
\bottomrule
\end{tabular}
}
\end{table}

\subsection{Audio Retrieval}
\label{subsec:audio_retrieval}

We introduce a new benchmark for audio hashing, spanning music, environmental sound, and speech domains. An overview of the datasets is provided in Table~\ref{tab:audiodatasets}.

The benchmark includes: \textbf{GTZAN}~\cite{gtzan} (music genres), \textbf{ESC-50}~\cite{esc50} (environmental sounds), \textbf{CREMA-D}~\cite{cremad} (speech emotions), and \textbf{VocalSound}~\cite{vocalsound} (human vocalizations). Due to the small size of these datasets compared to standard image benchmarks, we follow a setup similar to CIFAR-10, treating the train and database sets as identical.

We evaluate three state-of-the-art encoders: \textbf{CED}~\cite{ced}, a knowledge distillation framework that ensembles teacher models for efficient audio tagging; \textbf{Dasheng}~\cite{dasheng}, a large-scale self-supervised masked audio encoder trained on diverse audio for general classification; and \textbf{LAION-CLAP}~\cite{laionclap}, a multimodal contrastive learning model aligning audio and text pairs. Retrieval results are summarized in Table~\ref{tab:audio_retrieval}, following the same evaluation protocol as in Section~\ref{subsec:image_retrieval}.

\noindent\textbf{Analysis.}  
Performance trends are consistent with those observed in the image setting: \textit{Hashing-Baseline} incurs a modest loss relative to PCA-reduced features, which decreases at higher bit lengths. \textbf{CLAP} generally achieves the best performance due to its broad multimodal training, while \textbf{CED} remains competitive on \textbf{ESC-50} and \textbf{GTZAN}. \textbf{Dasheng} underperforms, suggesting that pretraining by reconstruction may produce embeddings that are not directly suitable for retrieval.

\section{Conclusion \& Perspectives}

We introduced \textit{Hashing-Baseline}, a simple yet effective method that leverages strong pre-trained model embeddings with classical hashing techniques. By producing compact binary codes, it enables fast, memory-efficient, and low-complexity retrieval while remaining entirely training-free, and it achieves competitive performance across both image and audio domains.

There is still room for improvement in how binary embeddings capture the full richness of pre-trained representations. Future work could explore lightweight hashing via parameter-efficient fine-tuning, allowing pre-trained models to adapt to different bit constraints with minimal computational overhead.

End-to-end representation learning jointly optimized with hashing also presents an exciting opportunity. Co-training the feature extractor and hashing module could better align the latent space with quantization constraints, further enhancing retrieval accuracy while preserving efficiency.

Finally, extending \textit{Hashing-Baseline} to training-free cross-modal retrieval (e.g., audio--text, image--text) could unlock scalable, high-performance multimodal retrieval systems, maintaining the benefits of compact binary codes without the need for full end-to-end training.

\section{Acknowledgment}

This work was supported by the Pl@ntAgroEco project, funded by the "Agence Nationale de la Recherche" (ANR) under the France 2030 program, within the “Agroécologie et Numérique” initiative (reference ANR-22-PEAE-0009). The authors gratefully acknowledge this support.


\bibliographystyle{IEEEbib}
\bibliography{refs}

\end{document}